\def\year{2022}\relax
\documentclass[letterpaper]{article} 
\usepackage{aaai22}  
\usepackage{times}  
\usepackage{helvet}  
\usepackage{courier}  
\usepackage[hyphens]{url}  
\usepackage{graphicx} 
\usepackage{natbib}  
\usepackage{caption} 
\DeclareCaptionStyle{ruled}{labelfont=normalfont,labelsep=colon,strut=off} 
\usepackage{algorithm}
\usepackage{algorithmic}

\usepackage{newfloat}
\usepackage{listings}
\floatstyle{ruled}
\newfloat{listing}{tb}{lst}{}
\floatname{listing}{Listing}
\title{Causal Discovery in Hawkes Processes 
\\by Minimum Description Length}
\author{
    Amirkasra Jalaldoust,\textsuperscript{\rm 1,2}
    Kate\v rina Hlav\' a\v ckov\' a-Schindler, \textsuperscript{\rm 3,4}
    Claudia Plant, \textsuperscript{\rm 3,5}
}
\affiliations {
    \textsuperscript{\rm 1} Department of Computer Science, Columbia University, New York, USA \\
    \textsuperscript{\rm 2} Department of Mathematical Science, Sharif University of Technology, Tehran, Iran \\
    \textsuperscript{\rm 3} Faculty of Computer Science, University of Vienna, Vienna, Austria \\
    \textsuperscript{\rm 4} Institute of Computer Science,  Czech Academy of Sciences, Prague, Czech Republic \\
    \textsuperscript{\rm 5} ds:UniVie, University of Vienna, Vienna, Austria \\
    jalaldoust@cs.columbia.edu, 
    katerina.schindlerova@univie.ac.at,
    claudia.plant@univie.ac.at
    \\
    PREPRINT
}

\usepackage{color}
\usepackage{amsfonts} 
\usepackage{graphicx}
\usepackage{amsfonts}
\usepackage{color}
\usepackage{algorithm}
\usepackage{algorithmic}
\usepackage{array}
\usepackage{booktabs}
\usepackage{siunitx}
\usepackage{subcaption}

\usepackage{amsmath}

\usepackage{amsthm}

\usepackage{array, booktabs, makecell, multirow}

    \usepackage{arydshln}
\usepackage{siunitx}

\newtheorem{theorem}{Theorem}

\newcommand{\balpha}{\boldsymbol\alpha}

\newcommand{\btheta}{\boldsymbol\theta}
\newcommand{\bTheta}{\boldsymbol\Theta}
\newcommand{\bomega}{\boldsymbol\omega}
\newcommand{\bmu}{\boldsymbol\mu}
\newcommand{\bgamma}{\boldsymbol\gamma}
\newcommand{\bGamma}{\boldsymbol\Gamma}
\newcommand{\eeta}{\boldsymbol\eta}
\newcommand{\bbeta}{\boldsymbol \beta}

\newcommand{\bX}{\boldsymbol X}

\newcommand{\bx}{\boldsymbol x}
\newcommand{\by}{\boldsymbol y}
\newcommand{\bs}{\boldsymbol s}
\newcommand{\bz}{\boldsymbol z}

\usepackage{url}

\DeclareMathOperator*{\argmin}{arg\,min}

\begin{document}

\maketitle
\begin{abstract}
 
Hawkes processes are a special class of temporal point processes which exhibit a natural notion of causality, as occurrence of events in the past may increase the probability of events in the future. Discovery of the underlying influence network among the dimensions of multi-dimensional temporal processes
is of high importance in disciplines where a high-frequency data is to model, e.g. in  financial data or  in seismological  data. This paper approaches the problem of learning Granger-causal network in multi-dimensional Hawkes processes. 
We formulate this problem as a model selection task in which we follow the minimum description length (MDL) principle. Moreover, we propose a general algorithm for MDL-based inference using  a Monte-Carlo method and we use it for our causal discovery problem. 
We compare our algorithm  with the state-of-the-art baseline methods   on synthetic and real-world financial data.
The synthetic experiments demonstrate superiority of our method  in causal graph discovery  compared to the baseline methods
with respect to the size of the data. The results of experiments with the  G-7 bonds price data are consistent with the experts' knowledge.

\end{abstract}
\section{Introduction}\label{intro}

In many applications, one needs to deal with multi-dimensional sequential data of irregular or asynchronous nature occurring in a continuous time. The examples can 
be
preferences of users in 
in social networks, interactions of areas during
earthquakes and aftershocks
in geophysics \cite{veen} or  high-frequency signals in
financial data \cite{bacry2015}. 
These data can be seen as event sequences containing multiple event types and modeled by a multi-dimensional Hawkes process (MHP)\cite{hawkes1971},  also known as a self-exciting  temporal point process.  
The main advantage of using Hawkes processes over e.g. Poisson processes is  that they  permit  to  model  the  influence  of the past  events on the current behavior of the process,  due  to their  memory  property.
Eichler \cite{Eichler2015} introduced  the  graphical Granger causal model for MHP in the form of an autoregressive structure for the intensities of  each marginal process. 

\noindent Methods to discover Granger-causal graphs in MHP search for a solution of a variable selection problem, mostly in terms of an optimum of a corresponding objective function constructed for the MHP at hand \cite{zhou2013learning,xu,zhou2013}.
These methods usually show good performance in scenarios with "long" horizon $T$, i.e. when $T$ is  greater  by
several orders of magnitude than the dimension of MHP $p$, however in the opposite case of "short" horizon, they often suffer from over-fitting. We define short event sequence when the time horizon is at most of order $100 \cdot p$ for a $p$-dimensional event sequence.

\noindent In our paper we approach causal discovery in MHP by the  minimum description (MDL) principle  which was as a model selection method
 introduced in \cite{rissanen1989} and developed in \cite{gruenwald2020}. 
Although MDL has  been already applied to graphical Granger-causality in \cite{hlavackova2020_ECAI} for Gaussian time series with a significant precision superiority over baseline methods,  application of MDL to Hawkes processes is  more challenging due to the nature of the processes. 
The contributions of our paper are summarized as follows.
\begin{itemize}
    \item  We present a general procedure for practical estimation of a MDL objective function  based  on  Monte-Carlo (MC) methods for estimation of integrals.
\item Using  this procedure  we  construct  an  MDL  objective  function  for  inference of the Granger-causal graph for multi-dimensional Hawkes processes.
\item We evaluate the performance of our causal discovery algorithm on both synthetic and real-world data, and compare our results with the baseline methods. Our method demonstrates a significant superiority in causal discovery
for short event sequences.
\end{itemize}
The paper is organized as follows: The multi-dimensional Hawkes processes and the minimum description length principle are defined in Section~\ref{Sec:preliminaries}. Section~\ref{Sec:our method} presents our method to estimate a general MDL function. Causal discovery in Hawkes processes is proposed in Section~\ref{Sec:causal}. Related work is discussed in Section~\ref{Sec:related}. Section~\ref{Sec:exp} presents experiments and discussion and our conclusion is in Section~\ref{Sec:conclusion}. The Appendix contains a proof and experimental setup.

 
\section{Preliminaries}\label{Sec:preliminaries}
\subsection{Notation}
Scalar variables are denoted by regular letters,  multi-dimensional variables bold,  random variables by capital letters (e.g. $\bX$), the support of each random variable by the same calligraphic letter (e.g. $\cal X$) and any realization or point in the support is denoted by lower-case letter (e.g. $\bx$). 
Matrices are denoted by Greek letters, and for matrices such as $\bomega$, $\bomega_i$ denotes the $i$-th row. All vectors are column vectors.

\subsection{Temporal Point Processes}
A temporal point process is a random process which is used to model occurrence of events in time. Each realization is a list of events $\{t_i\}$ with $t_i \in [0,T]$ and T is called horizon. The interval $[0,T]$ denotes the time window  in  which  the process was observed. A temporal point process can be equivalently represented by a counting process $U$ where $U(t)$ for $t \in [0,T]$ is the number of events happened prior to time $t$. 
Figure~1a visualizes a realization of a three-dimensional temporal point process.
For a temporal point process one may define the conditional intensity function \begin{equation}
    \lambda(t) = \mathbb{E}[dU(t)|{\cal H}_t] =  \lim_{\Delta t \to 0} \mathbb{E}[U(t + \Delta t) - U(t)| {\cal H}_t],
\end{equation}
where ${\cal H}_t$ is called the filtration at time $t$ consisting of all events prior to time $t$.
A multi-dimensional temporal point process is a set of coupled temporal point processes and it can represented by a set of counting processes $\{U_i\}_{i=1}^p$, where $U_i(t)$ denotes the number of events in the $i$-th process prior to time $t$.
Similarly, the conditional intensity function for the $i$-th dimension is \begin{equation}
    \lambda_i(t) = \mathbb{E}[dU_i(t)|{\cal H}_t].
\end{equation}

 \begin{figure}
\centering
\begin{subfigure}[b]{0.5\textwidth}
\hspace{0.9cm}  \includegraphics[width=0.72\linewidth]{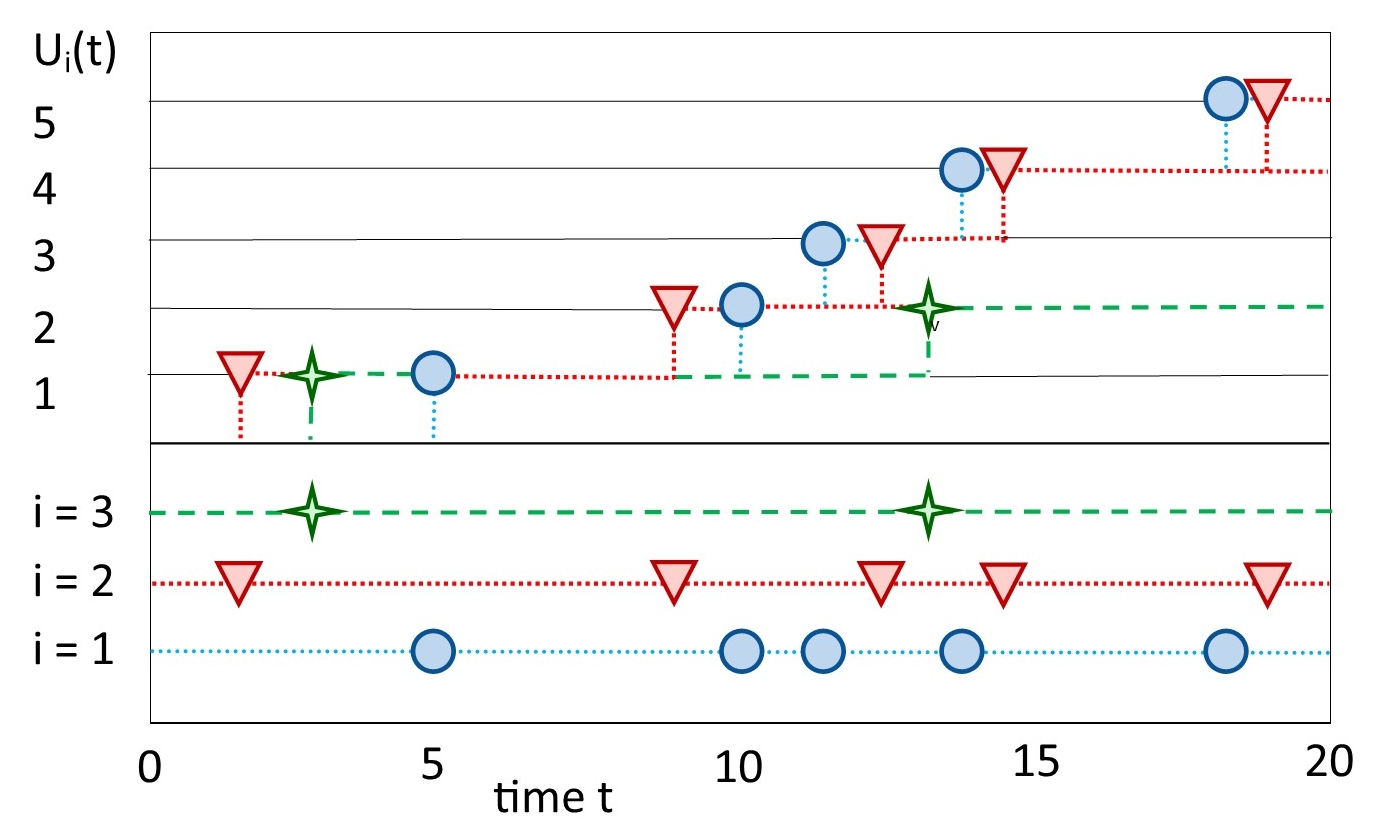}
   \caption{}
   \label{fig:Ng1} 
\end{subfigure}
\begin{subfigure}[b]{0.5\textwidth}
\hspace{0.9cm}
   \includegraphics[width=0.7\linewidth]{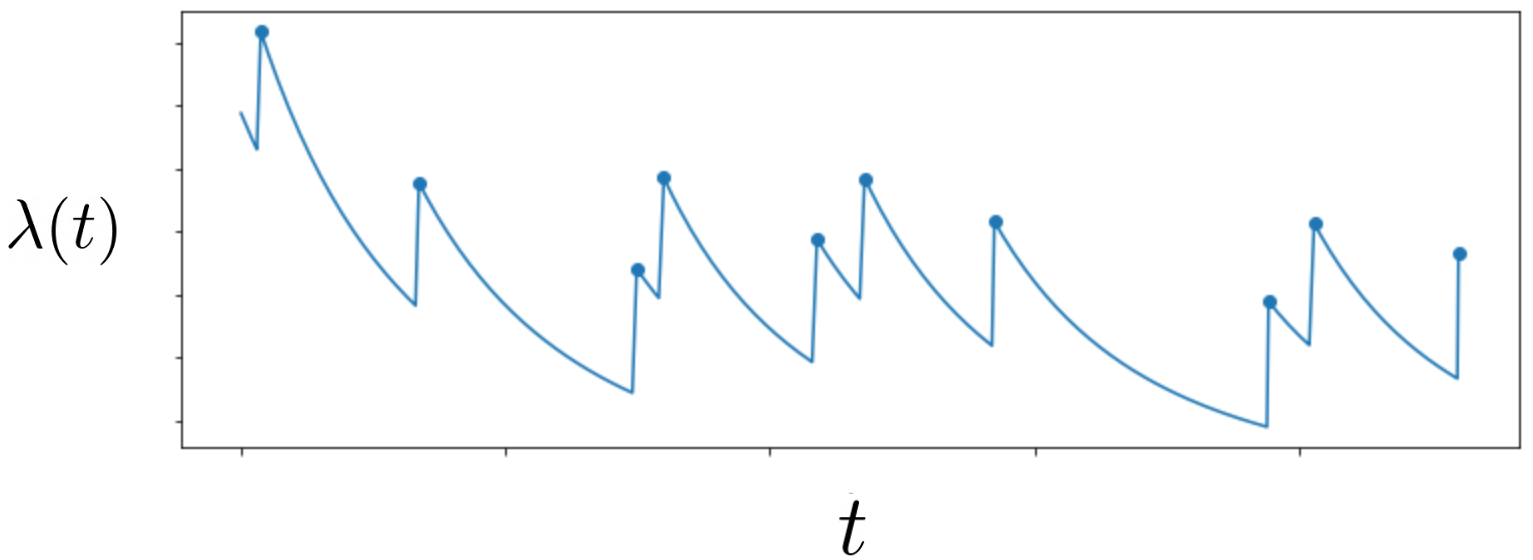}
   \caption{}
   \label{figure1}
\end{subfigure}

\caption[Two numerical solutions]{(a) Realization of a 3-dimensional temporal point process, and its corresponding counting process; (b) The intensity function of a 1-dim Hawkes process with exponential decay kernel.}
\end{figure}

\subsubsection{Granger-Causality in Temporal Point Processes} In multi-dimensional time-series variable $\bx_i$ Granger-causes variable $\bx_j$ when the future of $\bx_j$ is better predicted when taking into account the past of variable $\bx_i$ \cite{granger1969}. In multi-dimensional temporal point processes the events in the $j$-th dimension Granger-cause the events in the $i$-th dimension if \cite{kim2011granger}
\begin{equation}
    \lambda_i(t) = \mathbb{E}[dU_i(t)|{\cal H}_t] \neq \mathbb{E}[dU_i(t)|{\cal H}^{-j}_t],
\end{equation}
where ${\cal H}^{-j}_t$ is the history prior to time $t$ excluding events in the $j$-th dimension.

\subsubsection{Hawkes Processes}\label{tpp}
Introduced in \cite{hawkes1971}, a Hawkes process is a temporal point process with the intensity function following the equation
\begin{equation}\label{model_hp_single}
    \lambda(t) = \mu +
    \int_{-\infty}^{t} \phi(t-\tau) d U(\tau),
\end{equation}
where  $\mu >0$ is the exogenous baseline intensity, and $\phi: \mathbb{R}^+_0 \to \mathbb{R}^+_0$ is the kernel function, see e.g. Fig~1b. Similarly, a multi-dimensional Hawkes process (MHP) is defined to be a multi-dimensional temporal point process with conditional intensity function at each dimension $i=1,\dots, p$ following
\begin{equation}\label{MHP}
    \lambda_i(t) = \bmu_i +
    \sum_{j=1}^{p} \int_{-\infty}^{t} \phi_{ij}(t-\tau) d U_j(\tau).
\end{equation}

\subsubsection{Granger-Causality in MHP}

Eichler et al. in \cite{Eichler2015}
have  shown that the Granger-causality structure of MHP is fully
encoded in the corresponding kernel functions of the model. The result is stated in the following theorem.
\begin{theorem}[Eichler et al., 2017]\label{Eichler2015}
Let $U = \{U_i \}_{i=1}^p$ be a $p$-dimensional Hawkes
process with conditional intensity functions defined as in Eq. \ref{MHP}.
Then the events in the $j$-th dimension do not Granger-cause events in the $i$-th dimension if only if $\phi_{ij} \equiv 0$.
\end{theorem}

\subsection{Minimum Description Length}
The minimum description length (MDL),
introduced into statistical modeling by \cite{rissanen1989,barron1998} is a  principle based on compression of information. 
The most common version of the method as a model selection principle  makes use of two-part codes: the first part represents the information that one is trying to learn, such as the index of a model in a family of models (model selection) or parameter values (parameter estimation); the second part is an encoding of the data, given the model in the first part.

Statistical MDL learning is strongly connected to probability theory and statistics through the correspondence between codes and probability distributions.
While Bayesian approach is often useful in constructing efficient MDL codes, the MDL framework also accommodates other codes that have no assumption about the data-generating process. An example is the Shtarkov normalized maximum likelihood code \cite{shtarkov1978}, which we will use. 
Based on  \cite{gruenwald2020},
statistical models
(families of probability distributions) are of the form
 $M = \{p(.|{\btheta}) : \btheta \in  \bTheta \}$ parametrized by
a parameter space $\bTheta$ (usually a subset of a Euclidean space).
Further, we partition the parameter space $\bTheta$ into a family of disjoint restricted parameter subspaces $\{\bTheta_{\bgamma} : \bgamma \in \bGamma\}$ so that we can define families of models
$ \{M_{\bgamma}: \bgamma \in \bGamma\}$ where each
$M_{\bgamma} = \{ p(.|{\btheta}): \btheta \in \bTheta_{\bgamma}\}$ is a statistical model used to model data $\bx \in {\cal X}$, and each $p(.|{\btheta})$ represents a probability density function  on $\cal X$.

\subsubsection{Normalized Maximum Likelihood Distribution}
The normalized maximum likelihood (NML) distribution provides a general technique
to apply the minimum description length (MDL) principle to
statistical model selection.
In its general form, the NML distribution and the MDL estimators depend on a function
$v : \bTheta \to \mathbb{R}^+_0$ named luckiness function. The NML distribution for each model $\bgamma \in \bGamma$ is  given by
\begin{equation}\label{NML2}
p_{v|\bgamma}^{NML}({\bx}) = \frac{\max_{\btheta \in \bTheta_{\bgamma}} p({\bx}|{\btheta})v(\btheta)}
{\int_{\cal X} \max_{\btheta \in \bTheta_{\bgamma}} p({\bs}|{\btheta})v(\btheta) d{\bs}}.
\end{equation}
(It is  well-defined whenever the normalizing integral in (\ref{NML2}) is finite.) The logarithm of this integral,   called the model complexity  is   
\begin{equation}\label{comp2}
COMP(M_{\bgamma};v)= \log \int_{\cal X} \max_{\btheta \in \bTheta_{\bgamma}} p({\bs}|{\btheta})v(\btheta) d{\bs}.
\end{equation}

\noindent To simplify the notation, for each $\bx \in {\cal X}$ and $\btheta \in \bTheta$ let 
$R_v(\btheta;\bx) = p({\bx}|{\btheta})v(\btheta)$,
and let 
\begin{align}
r_v(\btheta;\bx) &  = \log R_v(\btheta;\bx) 
=\log p(\bx|{\btheta}) +  \log v(\btheta).
\end{align}
For any luckiness
function $v$, we define the MDL estimator based on $v$ for a specific model $M_{\bgamma} \subset M$ as
\begin{align}\label{MDLest-gamma}
\hat{\btheta}_{v|\bgamma}(\bx) = \argmin_{\btheta \in \bTheta_{\bgamma}} - r_v(\btheta;\bx).
\end{align}
Suppose now  a collection of models $M_{\bgamma}$ indexed by a finite $\bGamma$ and  specified luckiness functions $v_{\bgamma}$ on $\bTheta_{\bgamma}$, for each $\bgamma \in \bGamma$ and we pick a uniform distribution $\pi$ on $\bGamma$.
If we base the model selection on NML, we pick over $\bGamma$ the model minimizing  function
\begin{align} \label{L_v}
L_v(\bgamma;\bx) =& - \log\pi(\bgamma) -\log p^{NML}_{v|\bgamma}(\bx) \nonumber\\
=& -\log \pi(\bgamma) - r_v(\hat{\btheta}_{v|\bgamma}(\bx);\bx)\\
 &+ COMP(M_{\bgamma};v).\nonumber
\end{align}
The MDL function incorporates a trade-off between the goodness-of-fit measured by $r_v(\hat{\btheta}_{v|\bgamma}(\bx);\bx)$ and the model complexity measured by $COMP(M_{\bgamma};v)$.
Finally, the model selection based on MDL picks
\begin{equation}\label{MDL_model_selection}
\hat{\bgamma}^{MDL} = \argmin_{\bgamma \in \bGamma} L_v(\bgamma;\bx).
\end{equation}

\section{Our Method to Estimate the MDL Function}\label{Sec:our method}
We propose a new method to estimate the MDL objective function $L_v(\bgamma;\bx)$ for a given data $\bx \in {\cal X}$, a fixed model $M_{\bgamma} \subset M$, and an appropriate luckiness function $v$. We further apply this  general method to the problem of causal discovery in Hawkes processes.

\noindent According to Eq. \ref{L_v}, in order to estimate $L_v(\bgamma;\bx)$,  the terms $\log \pi(\bgamma)$, $r_v(\hat{\btheta}_{v|\bgamma}(\bx);\bx)$, and $COMP(M_{\bgamma};v)$ should be computed. As  $\pi(\bgamma)$ is known to us a priori, the first term $\log\pi(\bgamma)$ is computable.
We propose how to compute the other two terms - goodness-of-fit $r_v(\hat{\btheta}_{v|\bgamma}(\bx);\bx)$ and model complexity $COMP(M_{\bgamma};v)$.

\subsection{Computing the Goodness-of-Fit}\label{GOF}

Finding the MDL estimator $\hat{\btheta}_{v|\bgamma}(\bx)$ is the result of a minimization process over the restricted parameter space $\bTheta_{\bgamma}$. 
Convex optimization problems have a unique solution and there are efficient methods to find it.
To this end, we require two conditions:
\begin{enumerate}
    \item The restricted parameter space $\bTheta_{\bgamma}$ is a convex set. 
   \item The objective function measuring the negative goodness-of-fit
    \begin{equation}\label{r_v}
    -r_v(\btheta;\bx) = -\log(p(\bx|\btheta)) -\log(v(\btheta))
    \end{equation}
   is a convex function.
\end{enumerate}
Condition 1 depends on an appropriate partitioning of the parameter space and can be fulfilled in many scenarios  (e.g. when $\bTheta_{\bgamma}$ is a Euclidean space or a positive cone of it). 
Condition 2 can be fulfilled by an appropriate choice of luckiness function $v$; 
the objective function in Eq. (\ref{r_v})  involves two terms: \begin{enumerate}
    \item [i.] The negative log-likelihood term $-\log p(\bx|\btheta)$.
    \item [ii.] The negative log-luckiness term $-\log v(\btheta)$.
\end{enumerate}
The term (i) depends on the problem in hand, however, in many statistical scenarios it is convex in the model parameter $\btheta$ (e.g. in linear regression models and exponential families). The term (ii) is convex in $\btheta$ for any log-concave choice of $v$. Therefore, the summation of the two convex terms results in a convex objective function.
In summary, to have a convex optimization problem for finding the MDL estimator $\hat{\btheta}_{v|\bgamma}(\bx)$, we require:
\begin{enumerate}
    \item[1.] 
    $\bTheta$  to be partitioned into convex subsets $\{\bTheta_{\bgamma} \}_{\bgamma \in \bGamma}$.
    \item[2.] 
    Log-likelihood to be a concave function.
    \item[3.]
     Luckiness $v$ to be selected as a log-concave function.
\end{enumerate}

\subsection{Estimating Model Complexity}
In this subsection we propose estimation of the model complexity by using Monte Carlo (MC) simulation for estimation of integrals.
According to  definition of $R_v$ 
and Eq. \ref{MDLest-gamma} model complexity  can be rewritten as
\begin{equation}
    COMP(M_{\bgamma};v) = \log \int_{\cal X} R_v(\hat{\btheta}_{v|\bgamma}(\bs); \bs) d\bs.
\end{equation}
As discussed in subsection \ref{GOF}, we can efficiently compute a unique MDL estimator $\hat{\btheta}_{v|\bgamma}(\bs)$ for any data $\bs \in {\cal X}$.
For any parameter $\bz \in \bTheta$, we have
\begin{align}
    COMP(M_{\bgamma};v) &=  \log \int_{\cal X} \frac{R_v(\hat{\btheta}_{v|\bgamma}(\bs); \bs)}{p(\bs|\btheta = \bz)} p(\bs|\btheta = \bz) d\bs \nonumber\\
    &= \log \mathbb{E}_{\bX \sim p(.|\btheta = z)}[\frac{R_v(\hat{\btheta}_{v|\bgamma}(\bX); \bX)}{p(\bX|\btheta = \bz)}] \nonumber\\
    &= \log \mathbb{E}_{\bX \sim p}[\frac{R_v(\hat{\btheta}_{v|\bgamma}(\bX); \bX)}{p(\bX|\btheta)} | \btheta = \bz].
\end{align}
Note that in order  to rewrite the integral as above, the term   $p(\bs|\btheta = \bz)$ should be positive for all data $\bs \in {\cal X}$.
The term inside the expectation is a  function of the parameter $\bz \in \bTheta$ and the random variable $\bX$, so for simplicity we define
\begin{align} \label{q}
    Q_{v|\bgamma}(\bs,\bz) & = \frac{R_v(\hat{\btheta}_{v|\bgamma}(\bs); \bs)}{p(\bs|\bz)} \nonumber\\
    &= \frac{p(\bs|\hat{\btheta}_{v|\bgamma}(\bs))v(\hat{\btheta}_{v|\bgamma}(\bs))}{p(\bs|\bz)}
\end{align}
so that we have
\begin{align}
    COMP(M_{\bgamma};v)
    &= \log \mathbb{E}_{\bX\sim p}[Q_{v|\bgamma}(\bX,\btheta)|\btheta = \bz].
\end{align}
We can also randomize $\bz$ and then take the expected value. As the general  MDL method does not impose any distribution for $\btheta$, we can assume any arbitrary full support distribution on $\bTheta$ and  take random samples of $\btheta$ from it. Therefore,
\begin{align}
COMP(M_{\bgamma};v) 
&= \log \mathbb{E}_{\btheta \sim p}[\mathbb{E}_{\bX\sim p}[Q_{v|\bgamma}(\bX,\btheta)|\btheta = \bz]] \nonumber \\
&= \log \mathbb{E}_{\bX,\btheta \sim p}[Q_{v|\bgamma}(\bX,\btheta)] \label{estimator}.
\end{align}
We estimate the last expectation by taking multiple joint samples of $\bX,\btheta$ (Line 1 and Line 3 of Algorithm \ref{alg:COMP}), and by computing $Q_{v|\bgamma}$ at these points (Line 5 of Algorithm \ref{alg:COMP}) we report the average value as an unbiased estimation of $COMP(M_{\bgamma};v)$. One can also compute confidence intervals for our estimation using sample variance of the random draws of $Q_{v|\bgamma}(\bX,\btheta)$. Our steps to estimate model complexity are summarized in  Algorithm \ref{alg:COMP}.  First, according to distribution $p(\btheta)$, we draw random parameters $\bz_1, \bz_2, \dots, \bz_N$ where $N$ is the number of MC simulations. Next, for each parameter $\bz_i$, we simulate the data $\bs_i$ according to the distribution $p(\bX|\btheta=\bz_i)$. Finally, computing the MDL estimator $\hat{\btheta}_{v|\bgamma}(\bs_i)$ enables us to compute 
\begin{equation}
    Q_i = \frac{p(\bs_i|\hat{\btheta}_{v|\bgamma}(\bs_i))v(\hat{\btheta}_{v|\bgamma}(\bs_i))}{ p(\bs_i|\bz_i)},
\end{equation} which is an i.i.d. sample of $Q_{v|\bgamma}(\bX,\btheta)$. According to Eq. \ref{estimator}, the mean value of random samples $Q_1, Q_2, \dots, Q_N$ is an unbiased estimator for model complexity. Furthermore, the accuracy of our estimation increases by increasing number of MC simulations $N$; In fact, by the central limit theorem, the variance of our estimator for model complexity linearly decreases with increasing $N$.

\begin{algorithm}[tb]
\caption{Estimate $COMP(M_{\bgamma};v)$}
\label{alg:COMP}
\textbf{Input}: Model $\bgamma$.\\
\textbf{Given}: Luckiness $v$, number of MC simulations $N$.\\
\textbf{Output}: Estimation of model complexity $
\hat{C}$.\\ 
\begin{algorithmic}[1] 
\STATE Draw i.i.d. parameters $\bz_1, \bz_2, \dots, \bz_N$ w.r.t. $p(\btheta)$ 
\FOR{$1 \leq i \leq N$}
\STATE Draw data $\bs_i$ w.r.t. $p(\bX|\btheta=\bz_i)$
\STATE $\hat{\btheta}_{v|\bgamma}(\bs_i) \gets \argmin_{\btheta \in \bTheta_{\bgamma}} - \log p(\bs_i|\btheta) - \log v(\btheta)$
\STATE $Q_i \gets p(\bs_i|\hat{\btheta}_{v|\bgamma}(\bs_i))v(\hat{\btheta}_{v|\bgamma}(\bs_i))/ p(\bs_i|\bz_i)$
\ENDFOR
\STATE $\hat{C} \gets \log ( \text{mean}(\{ Q_i\}_{i=1}^p)$) 
\STATE \textbf{return} $\hat{C}$ 
\end{algorithmic}
\end{algorithm}

\section{Causal Discovery in Hawkes Processes}\label{Sec:causal}
In this section we propose an MDL-based
algorithm for causal discovery in Hawkes processes. We restrict ourselves to MHP with exponential decay kernel functions (exp-MHP). This is a class of MHP with  kernels defined as 
\begin{equation}
    \phi_{ij}(t) = \alpha_{ij} \exp(-\beta_{ij}t),
\end{equation}
where $\balpha$ is a $p\times p$ matrix called the influence coefficient matrix with non-negative entries, and $\bbeta$ is a known constant $p\times p$ matrix called decay matrix with positive entries. Hence, the intensity function for the $i$-th dimension is
\begin{equation}
    \lambda_i(t) = \mu_i + \sum_{j=1}^p \int_{-\infty}^t \alpha_{ij} \exp(-\beta_{ij}(t-s)) dU_j(s),
\end{equation}
where $\bmu$ is a $p$-dimensional vector of non-negative baseline intensities. Exp-MHP with known decay matrix $\bbeta$ can be characterized by the pair $(\bmu,\balpha)$, and we denote this pair as the parameters.

We focus on exp-MHP in this paper, however, the methodology can be utilized in other parametric settings as well; e.g. in MHP with power-law kernels \cite{bacry2016}  and in MHP with kernels defined by a set of basis functions.
As mentioned above, causal discovery task is to identify the causal influence network among  variables. In case of Granger-causality in Hawkes processes, this network is a directed graph where each node corresponds to  a Hawkes process. Any directed graph of $p$ nodes can be expressed by its adjacency matrix $\eeta \in \{0, 1\}^{p \times p}$ where $\eta_{ij} = 1$ iff there exists an edge from $j$-th node to the $i$-th node. For the causal graph of a MHP we have $\eta_{ij} = 1$ iff events in the $j$-th dimension Granger-cause the events in the $i$-th dimension. Applying Theorem \ref{Eichler2015} for exp-MHP  it holds $\eta_{ij} = 0$ iff $\alpha_{ij} = 0$, and therefore, causal discovery in exp-MHP corresponds to identifying the sparsity pattern of the influence coefficient matrix $\balpha$.

\subsection{Parameter Learning in exp-MHP}
Before we construct our MDL function for causal discovery, we need to define parameters $\btheta$ and its space as well as to know the log-likelihood function w.r.t. $\btheta$.
The parameters of the exp-MHP model are the influence coefficient matrix $\balpha$ and the baseline vector $\bmu$. In the $i$-th dimension it is \begin{equation}\label{vectortheta}
    \btheta_i = (\mu_i, \balpha_i^{\top})^{\top} \in (\mathbb{R}^+_0)^{p+1},
\end{equation}
and we define the parameter vector of exp-MHP as
\begin{equation}
    \btheta = [\btheta_1^{\top}, \btheta_2^{\top}, \dots, \btheta_p^{\top}]^{\top} \in (\mathbb{R}^+_0)^{p+p^2}.
\end{equation}
Any realization of a MHP can be seen as the collection of event sequences $\bx = \{\bx_i\}_{i=1}^p$ where each event sequence $\bx_i = (t^i_1, t^i_2, \dots, t^i_{n_i})^{\top}$ denotes the times when the events occurred in the $i$-th dimension.
The multi-dimensional conditional intensity can be computed  in  interval $[0,T]$ based on realization $\bx$ and the parameter vector $\btheta$ \cite{daley}. Hence, the negative log-likelihood for $\btheta$ is \begin{align}
 - \log p(\bx | \btheta)
 &= \sum_{i=1}^p - \log  p(\bx|\btheta_i) \label{log_lik_decomposition} \\
 &=  \sum_{i=1}^p  \left( \int_{0}^T\lambda_i(s)ds 
-  \sum_{j=0}^{n_i} \log \lambda_i(t^i_j) \right) \nonumber
\end{align}
and in exp-MHP  
\cite{ozaki1979maximum}
\begin{multline}\label{log_lik_convex}
-\log p(\bx|\btheta_i) =  \mu_i T
  + \sum_{j=1}^p \frac{\alpha_{ij}}{\beta_{ij}} \sum_{k=1}^{n_j} [1-\exp(-\beta_{ij}(T-t^j_k))]\\
  - \sum_{l=1}^{n_i} \log[ \mu_i + \sum_{j=1}^p \alpha_{ij} \sum_{k: t^j_k < t^i_l} \exp(-\beta_{ij}(t^i_l - t^j_k))].
\end{multline}
The search space for finding $\btheta_i$ which minimizes Eq. \ref{log_lik_convex} is the positive cone $(\mathbb{R}^+_0)^{p+1}$, and the objective function is convex in $\btheta_i$ \cite{ogata1981}.
Therefore a unique solution exists and efficient algorithms are available, e.g. \cite{bacry2015}.
\subsection{Causal Discovery As Model Selection}\label{CD-MDL}
Let $\bGamma$ be the set of all binary $p \times p$ matrices. For each binary matrix $\bgamma \in \bGamma$ the model $M_{\bgamma}$ is the set of all $p$-dimensional Hawkes process models with their causal graph having adjacency matrix $\bgamma$. 
According to the definition of Granger-causality in exp-MHP, the restricted parameter space $\bTheta_{\bgamma}$ contains parameter vectors representing $\bmu,\balpha$ such that $\alpha_{ij} = 0$ iff $\gamma_{ij} = 0$. The baseline vector $\bmu$ has no influence on causal discovery and our proposed model selection formulation.

In this way the parameter space $\bTheta$ is partitioned into disjoint subsets $\{\bTheta_{\bgamma} \}_{\bgamma \in \bGamma}$ as $\bTheta_{\bgamma}$ and $\bTheta_{\bgamma'}$ do not intersect for any different $\bgamma,\bgamma' \in \bGamma$, because for any $\btheta \in \bTheta$ the sparsity pattern of the influence matrix $\balpha$ uniquely determines the causal graph $\bgamma$. Thus finding the true model $M_{\bgamma^*} \subset M$ is equivalent to finding  the true causal graph with adjacency matrix  $\bgamma^*$. 

\begin{figure}
\centering
\includegraphics[width=\linewidth]
{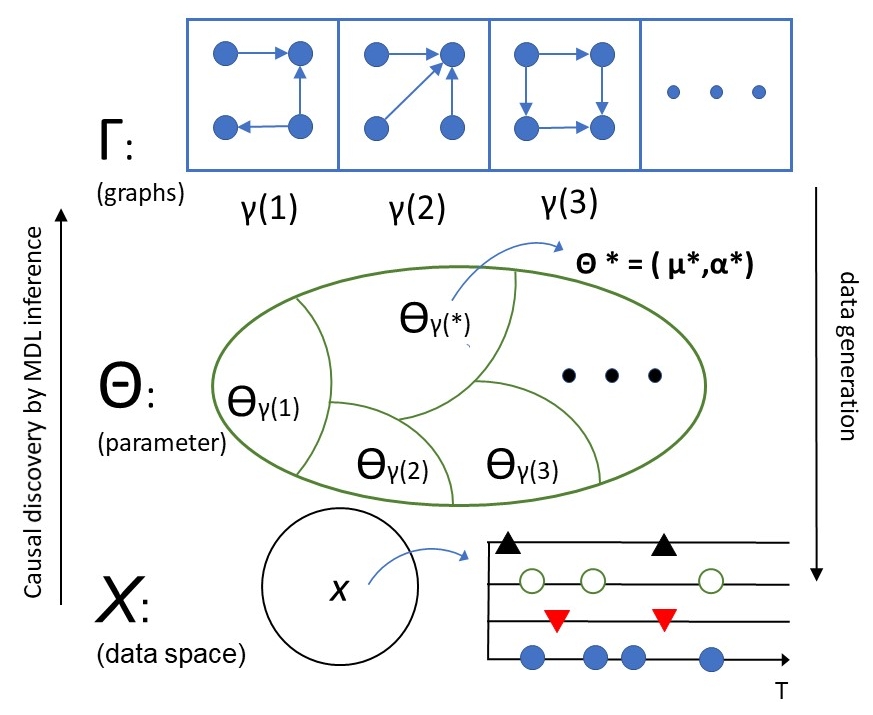}
\caption{The relationship between the true causal graph $\bgamma^{(*)}$, the true parameter $\btheta^*$, and the data $\bx$. The data generation mechanism (e.g., nature, right arrow) first decides the true causal graph $\bgamma^{(*)}$, then decides the parameter $\btheta^*$ (the excitation coefficients $\balpha^*$ and the exogenous intensities $\bmu$) in the restricted parameter space $\bTheta_{\bgamma^*}$, and finally draws a random realization (i.e., data) $\bx$ from this fully specified Hawkes process. Conversely, in the causal discovery task (left arrow) we are given the data $\bx$ and we use this hierarchy to infer the true causal graph.} 
\label{figure2}
\end{figure}

\subsubsection{MDL Objective Function for exp-MHP}\label{GOF-MHP}
We show that the MDL formulation for causal discovery discussed in subsection \ref{CD-MDL} satisfies the three conditions listed in the end of subsection \ref{GOF}. For any $\bgamma \in \bGamma$ containing $w >0$ non-zero entries,  matrix $\balpha$ has $w$ non-zero entries to be estimated, as well as p dimensions of the baseline vector $\bmu$ has to be estimated.
Hence, a total of $p+w$ non-negative parameters are to be estimated  and therefore, the restricted parameter space $\bTheta_{\bgamma}$ is the positive cone $(\mathbb{R}^+_0)^{p+w}$ which is a convex set (Condition 1). Further, according to Eq. \ref{log_lik_convex}, negative log-likelihood is a convex function (Condition 2). Thus, by choosing luckiness function $v$ to be log-concave (Condition 3) we get a convex optimization problem for finding the MDL estimator $$\hat{\btheta}_{v|\bgamma} = \argmin_{\btheta \in \bTheta_{\bgamma}} -\log p(\bx|\btheta) - \log v(\btheta).$$
For example, an appropriate uninformative choices for the luckiness function $v$ can be $v \equiv 1$, which we used in our experiments. Another example of appropriate choice which penalizes large baseline intensities and influence coefficients can be 
\begin{equation}\label{v_exp}
v(\btheta) = \prod_{i=1}^p \exp(-\mu_i) \prod_{i, j=1}^p \exp(-\alpha_{ij}).
\end{equation}
Efficient computation of the MDL estimator $\hat{\btheta}_{v|\bgamma}$ enables us to estimate model complexity $COMP(M_{\bgamma};v)$ according to Algorithm \ref{alg:COMP},
and thus we may estimate 
the MDL objective $L_v(\bgamma;\bx)$ as in Eq. \ref{L_v} for any specific causal graph $\bgamma \in \bGamma$.

\subsection{Causal Discovery Algorithm}
According to MDL criterion for model selection, to find the best model for describing the data $\bx$ and equivalently learning the underlying causal graph, we should find $\hat{\bgamma}^{MDL}$ from  Eq. \ref{MDL_model_selection}.
This can by found by the exhaustive search over $\bGamma$, or by   heuristics based on genetic algorithms, as e.g. in \cite{hlavackova2020}.
Figure~\ref{figure2} illustrates the connection of the mechanisms of  causal discovery by MDL and of data generation.

\noindent By imposing mild conditions  on the luckiness function $v$ and distribution  $\pi$ of the  models indexed in set $\bGamma$ as in Theorem~\ref{thm:decompose}, we reduce the complexity of the above search problem for finding $\hat{\bgamma}^{MDL}$.

\begin{theorem}\label{thm:decompose}
If in MDL-based model selection for exp-MHP,  
\begin{align} \label{thm:decompose_conditions}
    \pi(\bgamma) = \prod_{i=1}^p \pi_i(\bgamma_i), \quad 
    v(\btheta) = \prod_{i=1}^p v_i(\btheta_i),
\end{align}
then the MDL function can be rewritten as $p$ independent terms
\begin{equation} \label{decompose_L_v}
    L_v(\bgamma;\bx) = \sum_{i=1}^p L^i_v(\bgamma_i;\bx),
\end{equation}
such that each $L^i_v(\bgamma_i,\bx)$ can be computed by Algorithm \ref{alg:L_v}.

\end{theorem}
\noindent According to Theorem \ref{thm:decompose}, if the distributions over the causal graphs and the luckiness function are independent for each dimension of the Hawkes process, then we may optimize the MDL function for each of the dimensions separately, which enables us to perform parallel computation. The proof follows from the functional form of the MDL function, and it is provided in the Appendix. 
Algorithm 2 computes estimate of $L_v^i(\gamma_i; \bx)$ based on $COMP$ from Algorithm 1 and on input values $\bx$, $\bgamma$, $\pi_i, v_i$ and number of MC simulations.

\noindent By solving the independent optimization problems 
\begin{equation}
    \hat{\bgamma}_i^{MDL} = \argmin_{\bgamma_i \in \{0,1\}^p} L^i_v(\bgamma_i;\bx),
\end{equation}
 we get
\begin{equation}
    \hat{\bgamma}^{MDL} = [\hat{\bgamma}^{MDL}_1| \hat{\bgamma}^{MDL}_2| \dots| \hat{\bgamma}^{MDL}_p]^T
\end{equation}
as the solution to the optimization problem in Eq. \ref{MDL_model_selection}.

\noindent Our causal discovery method, summarised in 
Algorithm \ref{alg:causal_discovery}, executes the following steps: First, we capture $p$ as the number of coordinates of the data. Next, for every index $1\leq i\leq p$, we search exhaustively over the model space $\bGamma_i$ to find $\hat{\bgamma}_i$ which minimizes our estimation of the MDL objective function $L_v^i(\bgamma_i;\bx)$ computed by Algorithm \ref{alg:L_v}. By default, the model space $\bGamma_i$ is the set of all $p$-dimensional binary vectors. We call Algorithm \ref{alg:causal_discovery} ``MDLH" for MDL-based causal discovery in Hawkes processes.
\begin{algorithm}[tb]
\caption{Estimate $L^i_v(\bgamma_i;\bx)$}
\label{alg:L_v}
\textbf{Input}: The data $\bx$, model $\bgamma$, index $i$.\\
\textbf{Given}:  Distribution $\pi_i$, luckiness $v_i$, number of MC simulations $N$.\\
\textbf{Output}: Estimation of the MDL function $\hat{L}_i$\\

\begin{algorithmic}[1] 
\STATE $\hat{\btheta}_i \gets \argmin_{\btheta_i \in \bTheta_{\bgamma_i}} - \log p(\bx|\btheta_i) - \log v_i(\btheta_i)$
\STATE $\hat{C}_i
\gets$ estimate $COMP(M_{\bgamma_i};v)$ by Alg.~1
\STATE $\hat{L}_i \gets -\log \pi_i(\bgamma_i)
- \log p(\bx|\hat{\btheta}_i) -\log v_i(\hat{\btheta}_i)
+ \hat{C}_i$
\STATE \textbf{return} $\hat{L}_i$ 
\end{algorithmic}
\end{algorithm}
\begin{algorithm}[tb]
\caption{Causal Discovery by MDL}
\label{alg:causal_discovery}
\textbf{Input}: The data $\bx$\\
\textbf{Given}: Distribution $\pi$, luckiness $v$, number of MC simulations $N$, model spaces $\{\bGamma_i\}_{i=1}^p$ - default: $\bGamma_i = \{0,1\}^p$\\
\textbf{Output}: Inferred causal graph  $\hat{\bgamma}$ 
\begin{algorithmic}[1] 
\STATE $p \gets \text{dimension}(\bx)$
\FOR{$i \in \{1, 2, \dots, p\}$}
\STATE $\hat{\bgamma}_i \gets \argmin_{\bgamma_i \in \bGamma_i}$ estimate $L^i_v(\bgamma_i;\bx)$
 by Alg.~2
\ENDFOR
\STATE $\hat{\bgamma} \gets [\hat{\bgamma}_1| \hat{\bgamma}_{2}| \dots| \hat{\bgamma}_{p}] $ \hfill \texttt{//adjacency matrix}
\STATE \textbf{return} $\hat{\bgamma}$
\end{algorithmic}
\end{algorithm}

\subsubsection{Computational Complexity}
Our causal discovery method concluded in Algorithm \ref{alg:causal_discovery}
comprises of $p$ optimizations, each solved by $2^p$ calls of estimating MDL function by Algorithm \ref{alg:L_v}. 
The latter consists of i) parameter learning for the data $\bx$ and ii) a call of estimating model complexity by Algorithm \ref{alg:COMP} during which we run $N$ MC simulation by generating $N$ random data and learning the parameter for each of them. Therefore, each call of estimating MDL function by Algorithm \ref{alg:L_v} performs a total of $N+1$ parameter learning procedures. The computational complexity of these parameter learning procedures depends on the number of parameters to be learned (for $\bgamma_i$ with $w$ non-zero entries, the number of parameters is $w+1$) and the total number of observed events, however, for simplicity reasons we can  assume they 
have the same computational complexity. In conclusion,  to search over all causal graphs, we must perform $(N+1)\cdot p \cdot 2^p$  parameter learning operations under our causal discovery method.

\subsubsection{Sparse Causal Graphs} \label{sparse}

In scenarios when the expert knowledge  suggests a small upper-bound on the number of causes for each variable, we may reduce the computational complexity of our algorithm significantly. If the degree of each of the nodes in the causal graph is bounded by a constant $m \ll p$, we can achieve polynomial computational complexity. I.e., the model space $\bGamma_i$ for the $i$-th dimension  would  contain only $p$-dimensional binary vectors with at most $m$ non-zero entries. Thus, we have
\begin{equation}
| \bGamma_i | = \sum_{k=0}^m \binom{p}{k} < m\binom{p}{m} = {\cal O}(p^m).
\end{equation}
Therefore, we would have to perform ${\cal O}((N+1)\cdot p^{m+1})$ parameter learning procedures, which is polynomial in $p$ for constant $m \ll p$. We stress here however that the contribution of our work is not  in computational complexity but in the methodology and precision of causal discovery. 

\subsubsection{Amortization}  In situations when the problem specification (dimension $p$, decay matrix $\bbeta$, horizon $T$, model index distribution $\pi$, luckiness $v$, number of MC simulations $N$) is fixed, we can amortize most of the computational cost of our causal discovery method as follows: We compute and store the model complexities  as in Algorithm \ref{alg:COMP} for all possible inputs which costs $N\cdot p\cdot 2^p$ times of performing parameter learning procedure in the default scenario ($N\cdot p^{m+1}$ times in sparse graphs).
Next, for each query data we may run the causal discovery method in a total of $p\cdot2^p$ times of performing parameter learning procedures in Line 1 of Algorithm \ref{alg:L_v} ($p^{m+1}$ times in sparse graphs), while we use the pre-stored values for model complexities.

\subsubsection{Parallelization} 

  $p$ optimizations in Line 3 of Algorithm \ref{alg:causal_discovery} can be performed in parallel, as they are independent. Each  optimization consists of $2^p$ calls of Algorithm \ref{alg:L_v} in default scenario ($p^{m}$ calls in sparse graphs) which comprises $N+1$ independent parameter learning procedures. Therefore, all $(N+1)\cdot p\cdot 2^p$ steps ($(N+1)\cdot p^{m+1}$ in sparse graphs) are independent and can be performed in parallel.

\section{Related Work}\label{Sec:related}

\noindent We address here   the related work   on discovery of Granger-causal networks in MHP and on compression schemes related to causal discovery in general.
The method ADM4 in \cite{zhou2013learning} performs  variable selection  by using lasso and nuclear norm regularization simultaneously on the parameters to cluster variables as well as to obtain sparsity of the network. 
To detect a Granger-causal graph in MHP, \cite{xu} applied an EM algorithm  based on a penalized likelihood objective leading to temporal and group sparsity. 
The method NPHC \cite{achab2017} takes a non-parametric approach in learning the norm of the kernel functions to address the causal discovery problem. A moment-matching method  is used  fitting the second-order and third-order integrated cumulants of the process. NPHC is the most recent development in the literature and outperforms the state-of-the-art methods in many aspects.
These methods show good performance in scenarios with ``long" horizon T, however, in the opposite case of “short” horizon, they often suffer from over-fitting.

The MDL principle was applied to bivariate causal inference problem in \cite{marx2017,marx2018,budhathoki2018}. The considered causal inference  is however not the Granger one, as they discuss i.i.d. setting in contrast to sequentially ordered data. More recently, \cite{mian2021} extended these works to causal inference into multi-dimensional case using a greedy algorithm based on forward and backward search. The conditions on the convexity of the constructed  criterion functions is hard to impose and  the proposed greedy algorithm  does not guarantee to find a unique solution.
The MDL principle in the graphical Granger-causal models for Gaussian time-series was applied in \cite{hlavackova2020_ECAI}. 
A similar principle, connecting the idea of  compression of information with Bayesian inference, called the minimum message length, has been applied to graphical Granger models for Poisson time series in \cite{hlavackova2020_ECML} and for time series from exponential family in \cite{hlavackova2020}.
To our best knowledge, no work on causal discovery in MHP based on MDL has been published yet.

\section{Experiments and Discussion}\label{Sec:exp}
As mentioned before, we call the instance of our general method applied to MHP by MDLH.
We evaluate its performance  on both synthetic data with known ground truth, and real-world  data.  MDLH is implemented in python, and the experiments are performed on a   one Core Intel Xeon machine with 16 GB RAM. Our implementation and all experimental data are available at \url{https://dm.cs.univie.ac.at/research/downloads/} and \url{https://github.com/Amirkasraj/HawkesMDL}

\subsection{Synthetic Data}

\subsubsection{Baseline Methods} We consider i) NPHC \cite{achab2017} as a state-of-the-art method for causal discovery in MHP  since it  outperformed many of previous rival methods; ii) ADM4 \cite{zhou2013} has the same assumptions as ours and does model selection by a mix of lasso and nuclear regularization; iii) Information-theory based criteria for model selection including AIC, BIC, and HQ \cite{chen}; We refer the method with  the best score out of them as IC; iv) Regularized maximum likelihood (ML) and v) regularized least-squares (LS) with lasso, ridge, and elastic net regularizations.
Implementation of all these methods is provided in Python by using Tick package \cite{bacry2017tick}. 
For each of the comparison methods we report its best score among all possible regularizations and over a grid of hyperparameters. Each estimation method reports a matrix of kernel norms and we set threshold 0.01 to determine the causal graph from the real valued kernel norms.

\subsubsection{Data Generation Process}
As in Figure \ref{figure2}, we generate synthetic data. In all experiments, we allow for self-excitation in all dimensions, i.e., the diagonal entries of all adjacency matrices are non-zero. In default scenario (low dimensions), each non-diagonal entry of the adjacency matrix of the causal graph is randomly drawn from Bernoulli$(r)$. In sparse graphs (high dimensions), for each dimension, we draw the number of causes (other dimensions which affect this dimension) from $unif(\{0,1,\dots, m\})$, and then we uniformly chose a candidate from all possible subsets of other variables with that size.
For $p=7$ we go with default scenario with $r=0.3$, and for 
$p=20$ we go with sparse graphs scenario with $m=1$. Next, each entry of matrix $\balpha$ is drawn from unif$([0.1,0.2])$ and each entry of vector $\bmu$ is drawn from unif$([0.5,1.0])$. In our experiments, the decay matrix $\bbeta$ is the matrix of ones. 

\subsubsection{Evaluation and Results} 
We use F1 measure to evaluate the methods. F1 is suitable to evaluate the accuracy of estimated directed graphs represented by adjacency matrices, since gives the same importance to causal and non-causal connections.
Due to  the limitation of our computation resources,  the number of MC  simulations $N$ was 1000 in our algorithm  in all cases;  
 We only consider luckiness function $v \equiv 1$ and uniform distribution $\pi$. In each experimental setting we randomly generate 100 data generation processes and draw one random sample from each of them. 
As Table~1 demonstrates, our method  significantly outperforms all  baseline methods on short horizon.
The total failure of IC methods (an empty graph in all cases) is presumably  due their weak performance for short data.

\begin{table}[ht]\label{exp:t1}
    \centering
    \caption{Performance of MDL and  baselines in F1.} 
    \begin{tabular}{lllllll}
        \toprule p  & \multicolumn{3}{c}{$7$} & \multicolumn{3}{c}{$20$}\\
        \cmidrule(lll){2-4} \cmidrule(lll){5-7}
        T  & $200$ & $400$  & $700$   & $500$ & $1300$ & $2000$  \\
        \midrule
        MDLH
        &
        \textbf{77.4}  & \textbf{84.7}  & \textbf{89.3}  & \textbf{79.4}  & \textbf{82.8}  & \textbf{84.4}  \\
        \hline \\
        ADM4
        & 68.4 & 72.6  & 78.5  & 26.8  & 29.9  & 31.5  \\
        NPHC
        & 49.3  & 58.8  &   61.3 & 27.3 & 34.5  &  40.0 \\
        ML
        & 68.7  & 74.6  & 80.4  & 25.8  & 28.2  & 29.4  \\
        LS
        & 68.3  & 74.4  & 76.9  & 26.4  & 29.8  & 31.3  \\
        IC
        & NA  & NA  & NA  & NA  & NA  & NA  \\
        Random
        & 30.0  & 30.0  & 30.0  & 7.5  & 7.5  & 7.5 \\
        \bottomrule
    \end{tabular}
\end{table}

\subsection{Real-World Data} 

\subsubsection{G-7 Bonds} We use daily return volatility of sovereign bonds of 7 large and developed economies called G-7 including USA, Germany, France, Japan, UK, Canada, Italy from 2003-2014 as in \cite{demirer2018}. The goal is to discover the underlying influence network among the sovereign bonds. 

\subsubsection{Shock Identification} As the data is a time-series  and not a point process, to identify shocks (events in point process) in the daily return volatility, we roll a one year window over the data, and in each dimension if the latest value of the window is among the top 20 percent in the rolling window we register an event in that dimension for that day. The number of events registered in each dimension is around 500, and  applying the  knowledge gained in the synthetic experiments with a similar scenario,
we assumed this data is an instance of our synthetic data with $T=400$.
\subsubsection{Results} 
\begin{figure}
 \centering
 \includegraphics[width=0.6\linewidth]
 {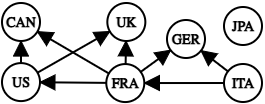}
 \caption{G-7 causal graph inferred by MDLH}
 \label{figure3}
 \end{figure}

Only MDLH and ML were tested on the G-7 data,  since ML outperfomed other baseline methods for $p=7$ in the experiments with synthetic data. MDLH discovers the graph depicted in Figure \ref{figure3}. (The self-loops of all nodes  were omitted). The structure is plausible when considering the network discovered by \cite{demirer2018}. Also, the graph corresponds to the expert knowledge from the domain, e.g., that Japanese bond neither influences nor gets influenced by other G-7 countries. Moreover, the influence of French bonds on the US, UK and other big economies can be affected  by the fact that  France accused  U.S of 'economic war' in 2003 which spread in most of the world media \cite{CNN2003}.
On the other hand, ML discovers bi-directed edge between US and Japan which contradicts to the expert knowledge given in 
\cite{demirer2018}.
To conclude, the MDLH gives a more plausible graph than ML.

\section{Conclusion}\label{Sec:conclusion}
We presented a general procedure for practical estimation of an MDL-based  objective function  using Monte-Carlo  integral  estimation.
 In this procedure  we  constructed  an  MDL  objective  function  for  inference of the Granger-causal graph for multi-dimensional Hawkes processes.
We demonstrated significant superiority  of our method in causal discovery
for short event sequences in synthetic experiments. In a real experiment with G-7 bonds,  our method gives a more plausible causal graph than  the baseline method.

\section*{Appendix}
\subsection{Proof of Theorem 2}

 \begin{proof}
 Define
 \begin{align}
     \hat{\btheta}_{v|\bgamma_i}(\bx) = \argmin_{\btheta_i \in \bTheta_{\bgamma_i}} - \log p(\bx|\btheta_i) - \log v_i(\btheta_i)\nonumber,
 \end{align}
 where $\bTheta_{\bgamma_i}$ is the space of all possible values for $\btheta_i$. We have
 \begin{align}
     \hat{\btheta}_{v|\bgamma}(\bx) &=
     \argmin_{\btheta \in \bTheta_{\bgamma}} - \log p(\bx|\btheta) - \log v(\btheta)\nonumber\\
     &=[\hat{\btheta}_{v|\bgamma_1}(\bx)^T, \hat{\btheta}_{v|\bgamma_2}(\bx)^T, \dots, \hat{\btheta}_{v|\bgamma_p}(\bx)^T]^T\nonumber.
 \end{align}
 Further we define
 \begin{equation}
 COMP(M_{\bgamma_i};v) = \log \int_{\cal X} p(\bs|\hat{\btheta}_{v|\bgamma_i}(\bs)) v_i(\hat{\btheta}_{v|\bgamma_i}(\bs)) d\bs\nonumber.
 \end{equation}
 
 We have
 \begin{align}
     COMP(M_{\bgamma};v) &= \log \int_{\cal X} p(\bs|\hat{\btheta}_{v|\bgamma}(\bs)) v(\hat{\btheta}_{v|\bgamma}(\bs)) d\bs\nonumber\\
     &= \log \int_{\cal X} [\prod_{i=1}^p p(\bs|\hat{\btheta}_{v|\bgamma_i}(\bs)) v_i(\hat{\btheta}_{v|\bgamma_i}(\bs))] d\bs\nonumber\\
    &=\log \prod_{i=1}^p \int_{\cal X} p(\bs|\hat{\btheta}_{v|\bgamma_i}(\bs)) v_i(\hat{\btheta}_{v|\bgamma_i}(\bs)) d\bs\nonumber\\
     &= \sum_{i=1}^p \log \int_{\cal X} p(\bs|\hat{\btheta}_{v|\bgamma_i}(\bs)) v_i(\hat{\btheta}_{v|\bgamma_i}(\bs)) d\bs\nonumber\\
     &= \sum_{i=1}^p COMP(M_{\bgamma_i};v)\nonumber.
 \end{align}
 As in Eq. 22  
 in the paper, the negative log-likelihood can be written in independent terms for each dimension. Therefore,
 for each dimension $1\leq i\leq p$ we can write
 \begin{align} \label{L_v^i}
 L_v^i(\bgamma_i;\bx) :=& -\log \pi_i(\bgamma_i)
  - \log p(\bx|\hat{\btheta}_{v|\bgamma_i}(\bx))\\
  &- \log v_i(\hat{\btheta}_{v|\bgamma_i}(\bx))
 + COMP(M_{\bgamma_i};v)\nonumber.
 \end{align}
 Hence, we have
 \begin{align}
     \sum_{i=1}^p L_v^i(\bgamma_i;\bx) =& - \sum_{i=1}^p \log \pi_i(\bgamma_i) - \sum_{i=1}^p \log p(\bx|\hat{\btheta}_{v|\bgamma_i}(\bx)) \nonumber\\
     & - \sum_{i=1}^p \log v_i(\hat{\btheta}_{v|\bgamma_i}(\bx)) \nonumber\\
     & + \sum_{i=1}^p COMP(M_{\bgamma_i};v) \\
     =& - \log \pi(\bgamma) - \log p(\bx|\hat{\btheta}_{v|\bgamma}(\bx)) \nonumber\\
     & - v(\hat{\btheta}_{v|\bgamma}(\bx)) + COMP(M_{\bgamma};v)\\
     =& L_v(\bgamma;\bx).
 \end{align}
Thus, $L_v^i$ as defined above satisfies Eq. \ref{decompose_L_v} as required. Algorithm~2 
 summarizes the procedure for computing $L_v^i(\bgamma_i;\bx)$. First, we compute the MDL estimator $\hat{\btheta}_i$ for the $i$-th dimension by optimizing the goodness-of-fit, which is a convex optimization problem for an appropriate choice of luckiness function $v$, as discussed in subsection  4.3.
 Next, we estimate the model complexity
 by using Algorithm~1. 
 Finally, we compute MDL objective as in Eq. \ref{decompose_L_v}.
 \end{proof}

\subsection{Experimental Setup}
The comparison methods are estimation methods which search for the  MHP kernel functions and baseline vector based on the data. To extract a causal graph from the achieved output, based on Theorem 1, we put a threshold on the kernel norm to distinguish zero and non-zero kernels. This threshold is set to 0.01.

Each of the comparison methods has a penalty and a  level of regularization given. For ML, LS, and NPHC we have penalties: L1 (lasso), L2, elastic net, and none. For ADM4 we have lasso-nuclear penalization. We evaluated each of the baseline methods with all possible penalties and for a set of possible values (levels) for regularization $$C \in \{1, 2, 5, 10, 20, 50, 100, 200, 500, 1000, 2000, 5000, ...\}.$$ 
For ADM4, we also used the nuclear lasso ratio; For each of the above settings we also considered the lasso-nuclear-ratio taking value in $\{ 0, 0.1, 05, 0.9, 1 \}$.

The reported numbers for ``Random" in Table 1 are the result of a random adjacency matrix with the same number of non-zero entries as the average test case. We report the highest F1 score achieved by each method based on different hyper-parameters. As we do not perform train/test validation and instead we take the highest F1, the validated  F1 scores for baseline methods would be presumably lower.

Information criteria (AIC, BIC, and HQ)  generally do not perform well for small data and it was the case also in our experiments. These methods  rely on reducing model loss (i.e., negative log-likelihood) at the price of adding new parameters to the model. The least price that these models suggest for increasing the size of parameter set is about 1 unit of log-likelihood, hence, these model selection methods allow for adding any edge to the graph only if the log-likelihood could be increased by at least 1 unit compared to the empty graph model. This is not the case for a  small data set ("short" data), as the value of log-likelihood that we have for the naive (empty) model and the maximum-likelihood model are both very small (of order 0.01 or 0.1 in all of our experimental settings). Therefore, the 1 unit improvement is not possible and this prevents the IC methods from discovering any edge in the causal graph.

Our method is an MDL-based model selection with no hyper-parameters. However, we can choose the number of MC simulations $N$  for integral estimation, and the higher N the better estimate. Limited by our computational resources, we used 1000 iterations for  the default case (for dimension $p=7$), and 500 iterations for the sparse graph scenario (for dimension $p=20$). As discussed in Section 4 in Subsection Amortization, we first do the MC simulations and compute model complexity values, which takes about one hour in each experimental setting (i.e.,  with fixed $p$ and $T$), and then we perform 100 test runs, each taking a about ten seconds.

\subsubsection{Real-world Data}
We observed n our synthetic experiments  that method  ML  outperformed the other baseline methods for $p=7$ and $T=400$.  Elastic net regularization was the best regularization for ML  also in our synthetic experiments. So we  ran this method on the data for the set of levels for regularization as listed above, and in all cases the bi-directed edge between US and Japan as an edge of the causal graph was returned. This  is not plausible since it contradicts the expert knowledge from the domain.

\section*{Acknowledgements} 
This
work was funded  in part by the Austrian Science Fund (FWF) I5113-N, 
in part  within the Austrian Climate and Research Programme ACRP project MEDEA KR19AC0K17614, 
and in part by the Czech Science 
Foundation project GA19-16066S. 

\bibliographystyle{aaai22}
\bibliography{main_paper}

\end{document}